\documentclass[letterpaper]{article} 
\usepackage{main}  
\usepackage{times}  
\usepackage{helvet}  
\usepackage{courier}  
\usepackage[hyphens]{url}  
\usepackage{graphicx} 
\usepackage{natbib}  
\usepackage{caption} 
\usepackage{amsmath}
\usepackage{xcolor}
\usepackage{booktabs}
\usepackage{booktabs} 
\usepackage{multirow}
\usepackage{algorithm}
\usepackage{algorithmic}
\usepackage{newfloat}
\usepackage{listings}
\usepackage{threeparttable}

\DeclareCaptionStyle{ruled}{labelfont=normalfont,labelsep=colon,strut=off} 
\floatstyle{ruled}
\newfloat{listing}{tb}{lst}{}
\floatname{listing}{Listing}
\nocopyright

\title{Synthesizing Attitudes, Predicting Actions (SAPA): Behavioral Theory-Guided LLMs for Ridesourcing Mode Choice Modeling}

\author{
    Mustafa Sameen\textsuperscript{\rm 1}, 
    Xiaojian Zhang\textsuperscript{\rm 1, 2},
    Xilei Zhao\textsuperscript{\rm 1},
}
\affiliations{
    \textsuperscript{\rm 1}University of Florida,
    \textsuperscript{\rm 2}Kittelson \& Associates\\

    \textbf{mustafasameen@ufl.edu}, 
    \textbf{xczhang@kittelson.com},
    \textbf{xilei.zhao@essie.ufl.edu}, 
    
}

\begin{document}

\maketitle

\begin{abstract}
Accurate modeling of ridesourcing mode choices is essential for designing and implementing effective traffic management policies for reducing congestion, improving mobility, and allocating resources more efficiently. Existing models for predicting ridesourcing mode choices often suffer from limited predictive accuracy due to their inability to capture key psychological factors, and are further challenged by severe class imbalance, as ridesourcing trips comprise only a small fraction of individuals’ daily travel.
 To address these limitations, this paper introduces the \textit{\textbf{S}ynthesizing \textbf{A}ttitudes, \textbf{P}redicting \textbf{A}ctions} (\textbf{SAPA}) framework, a hierarchical approach that uses Large Language Models (LLMs) to synthesize theory-grounded latent attitudes to predict ridesourcing choices. SAPA first uses an LLM to generate qualitative traveler personas from raw travel survey data and then trains a propensity‐score model on demographic and behavioral features, enriched by those personas, to produce an individual‐level score. Next, the LLM assigns quantitative scores to theory‐driven latent variables (e.g., time and cost sensitivity), and a final classifier integrates the propensity score, latent‐variable scores (with their interaction terms), and observable trip attributes to predict ridesourcing mode choice. Experiments on a large-scale, multi-year travel survey show that SAPA significantly outperforms state-of-the-art baselines, improving ridesourcing choice predictions by up to \textbf{75.9\%} in terms of PR-AUC on a held-out test set. This study provides a powerful tool for accurately predicting ridesourcing mode choices, and provides a methodology that is readily transferable to various applications.
\end{abstract}

\begin{figure}[t]
\centering
\includegraphics[width=.8\columnwidth]{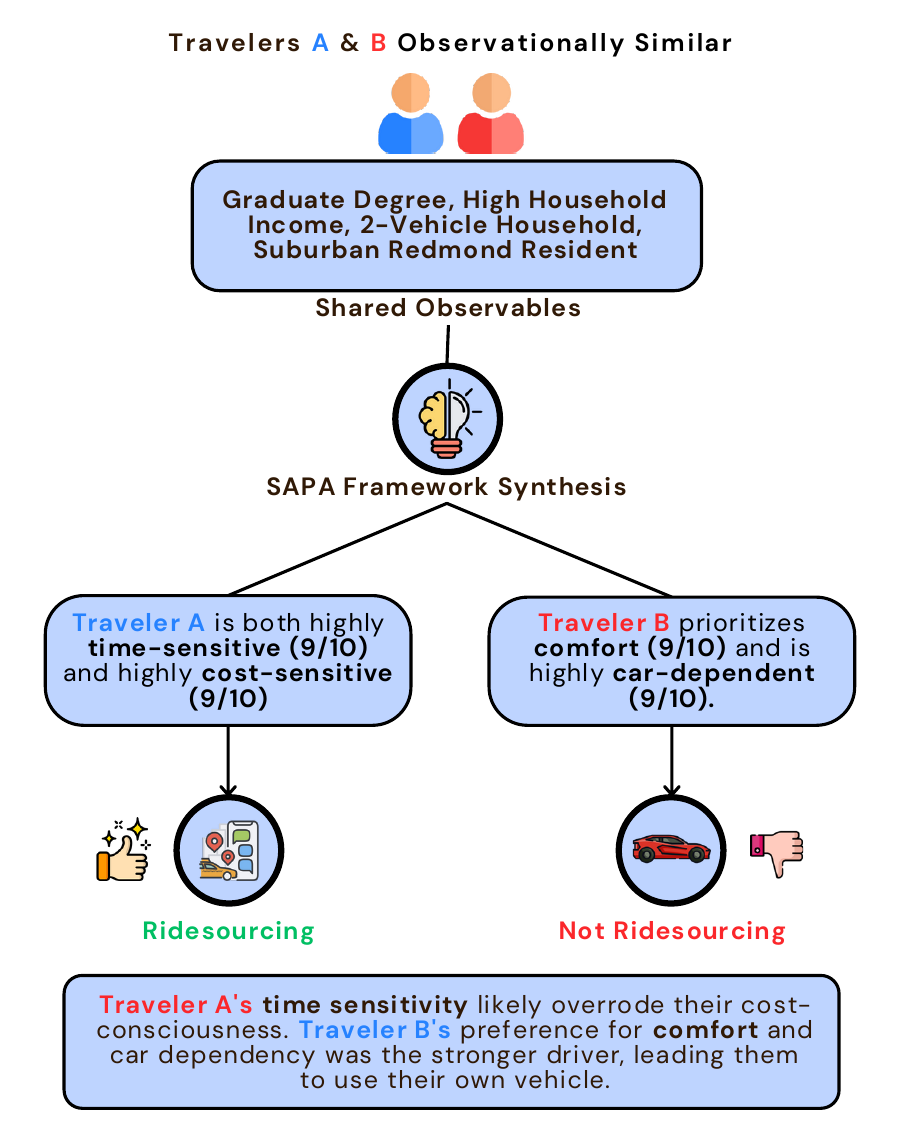}
\caption{SAPA uncovers latent attitudes that explain why two seemingly identical travelers make different real-world choices.}
\label{fig:intro_concept}
\end{figure}

\section{Introduction}

The proliferation of ridesourcing services like Uber and Lyft has fundamentally altered the urban mobility landscape, offering unprecedented convenience to millions of travelers \citep{mitropoulos2021factors, zhang2022machine}. However, this disruption brings complex challenges for transportation planning and policy, including concerns about increased traffic congestion, competition with public transit, and equitable access \citep{schaller2018, liang2023short}. Accurate ridesourcing models are crucial for designing sustainable and efficient urban transportation systems.

However, accurately modeling ridesourcing choices is hindered by several critical research gaps. First, large national/regional household travel surveys, while rich in sample size, almost universally lack psychometric data on the latent attitudes that are known to be powerful drivers of travel mode choices \citep{benakiva2002, walker2002,mokhtarian2024pursuing}, while smaller, specialized surveys that contain such data often lack scale and generalizability \citep{hess2018analysis}. Second, ridesourcing trips represent a rare event, typically accounting for only 1-2\% of total trips \citep{WuMacKenzie2021_evolution}. This severe class imbalance biases standard classification models, which struggle to learn the patterns of the minority class and often fail to predict the very behavior of interest \citep{hossain2023rare, he2023, cartus2023outcome, fan2023advancements}. The statistical challenge of learning from imbalanced data is well-documented \citep{he2009}, and methods to address it are less effective when key distinguishing features of the rare class, such as attitudes, are missing from the data in the first place.

While the inferential power of Large Language Models (LLMs) presents a tempting solution to these data challenges, a direct application is methodologically flawed. Simply using an LLM as a black-box predictor, though potentially accurate, would ignore decades of established behavioral science. Such a model would lack the interpretability required for policy analysis—it could predict \textit{what} a traveler might do, but not \textit{why}. This creates a critical methodological deadlock: a direct LLM application is atheoretical and opaque, rendering it unusable for actionable policy insights \citep{Goje2025_llm_interpretability, ThapaEtAl2025_css_llm_issues}.

The central challenge is therefore not \textit{if} LLMs can be used, but \textit{how} they can be integrated in a structured, theoretically-sound manner to produce trustworthy and actionable insights. This research addresses the overarching question: \textit{How can we construct a behaviorally rich and generalizable choice model for a rare event like ridesourcing when direct attitudinal data is unavailable?}

To bridge these gaps, this paper proposes the \textbf{Synthesizing Attitudes, Predicting Actions (SAPA)} framework, a novel hierarchical approach that uses behavioral theory-guided Large Language Models (LLMs) to enrich raw travel survey data with synthesized latent attitudes in order to accurately predict ridesourcing mode choices (see Figure \ref{fig:intro_concept}) \citep{argyle2024}. The SAPA framework first uses an LLM to process readily available demographic and spatial data for each traveler to generate a qualitative \textit{traveler persona}. Crucially, the synthesis process is not arbitrary. The framework's connection to established behavioral theory is operationalized in the second stage, where the LLM is prompted to produce quantitative scores for a predefined set of latent variables, such as \texttt{Time Sensitivity} and \texttt{Tech Affinity}, which are drawn directly from frameworks such as the Theory of Planned Behavior (TPB) \citep{ajzen1991, ajzen2020theory} and the Technology Acceptance Model (TAM) \citep{davis1989, davis1993user}. This methodology provides a scalable and cost-effective way to overcome the missing data problem, enabling the development of behaviorally sophisticated models using the vast archives of existing observational travel data \citep{long2024llms}.

Our experiments, validated on a large-scale, multi-year travel survey from the Puget Sound region, allow us to make the following significant contributions:
\begin{enumerate}
    \item \textbf{A Methodology for Behavioral Theory-Guided LLM-Powered Feature Synthesis:} We present and validate a structured methodology for using an LLM to synthesize theory-grounded latent attitudes from raw survey data, demonstrating that the resulting features are essential for model generalization and predictive accuracy.
    \item \textbf{A Novel Two-Stage Hierarchical Framework:} We embed this synthesis method within a robust two-stage framework that first models an individual's general propensity to use ridesourcing (Stage 1) and then predicts the trip-level choice (Stage 2). The robustness of the framework is demonstrated through rigorous validation on a large-scale travel survey.
    \item \textbf{Comprehensive Empirical Validation and Insights:} Through extensive ablation studies, we demonstrate that the SAPA framework consistently and significantly outperforms strong baselines across multiple classifiers. Our results highlight the framework's ability to uncover context-specific behavioral drivers and demonstrate its robustness, showing that the synthesized features provide value regardless of the final classifier choice.
\end{enumerate}

\section{Literature Review}

Our research is situated at the junction of three distinct but increasingly interconnected domains: behavioral modeling of travel mode choice, the application of machine learning to rare events, and the emerging use of Large Language Models (LLMs) for social simulation. This review synthesizes these fields to motivate our framework, which leverages the strengths of all three to address a long-standing data problem in transportation science.

\subsection{The Ridesourcing Choice Problem: Beyond Time and Cost}
The decision to use a ridesourcing service is a complex behavior influenced by a combination of situational factors, socio-demographics, and, critically, latent psychological attitudes \citep{mitropoulos2021factors}. While observable trip characteristics like travel time and cost are primary drivers \citep{ashkrof2023relocation}, their influence is mediated by \textbf{unobserved heterogeneity}, such as an individual's value of time (VoT) \citep{rose2025what}.The challenge of incorporating these latent factors has traditionally been the domain of econometric models, particularly Hybrid Choice Models (HCMs), which formally integrate latent variables into discrete choice frameworks \citep{benakiva2002, walker2002, hess2018analysis}. However, the demanding data requirements of HCMs - large travel surveys with attitudinal data - have limited their application to observational datasets \citep{yanez-manzilla2021effect}. Our work offers a new path forward by using machine learning classifiers and an LLM to quantitative scoring the missing latent variables, thereby outmaneuvering the need for specialized surveys while still capturing behavioral richness.

\subsection{Theoretical Foundations for Synthesizing Attitudes}
To ensure our synthesized latent variables are behaviorally meaningful, we ground their definitions in foundational psychological theories. The \textbf{Theory of Planned Behavior (TPB)}, which posits that behavioral intention is shaped by attitudes and perceived control, provides a basis for key variables like \texttt{Time\_Sensitivity} and \texttt{Cost\_Sensitivity} \citep{ajzen1991, ajzen2020theory, ciocirlan2025road}. Similarly, the \textbf{Technology Acceptance Model (TAM)}, which links perceived usefulness and ease of use to technology adoption, informs our \texttt{Tech\_Affinity} variable, capturing dispositions toward the applications that enable ridesourcing services \citep{davis1989, davis1993user, gundogan2024tam}. By grounding our LLM's inference task in these well-validated theories, we impose a behaviorally plausible structure on the synthesis process.

\subsection{Machine Learning for Rare Events and Imbalanced Data}
From a machine learning perspective, predicting ridesourcing choice is a classic rare event problem. The resulting \textbf{class imbalance} poses a significant challenge for standard classification algorithms, which are often biased toward the majority class and can achieve high accuracy simply by ignoring the rare event \citep{hossain2023rare, he2009, cartus2023outcome, fan2023advancements}. While data-level approaches like the Synthetic Minority Over-sampling Technique (SMOTE) can rebalance training data \citep{chawla2002}, their effectiveness is limited when the underlying behavioral signals that distinguish the rare class are missing from the data in the first place. Our framework incorporates SMOTE in the Stage 1 propensity score model to directly address this imbalance issue during feature engineering.

\subsection{The New Synthesis: LLMs for Behavioral Modeling}
Our work is at the forefront of a new paradigm that leverages the inferential capabilities of LLMs for social science research, often through \textbf{persona-driven prompting} where an LLM adopts a specific role to generate contextually appropriate outputs \citep{argyle2024, ge2024persona, long2024llms}. However, this emerging methodology is not without challenges, as researchers have raised serious concerns about the reliability and potential biases of LLM-generated data \citep{veselovsky2024synthetic, veselovsky2023}. Our framework is designed as a direct response to these concerns. By grounding the LLM's inference task in established theory and using its output as features in a downstream classification model, we use predictive performance as a validation mechanism, providing a structured approach that leverages the promise of LLMs while guarding against their known pitfalls.

\section{SAPA Framework}
\begin{figure*}[t]
    \centering
    \includegraphics[width=1.87\columnwidth]{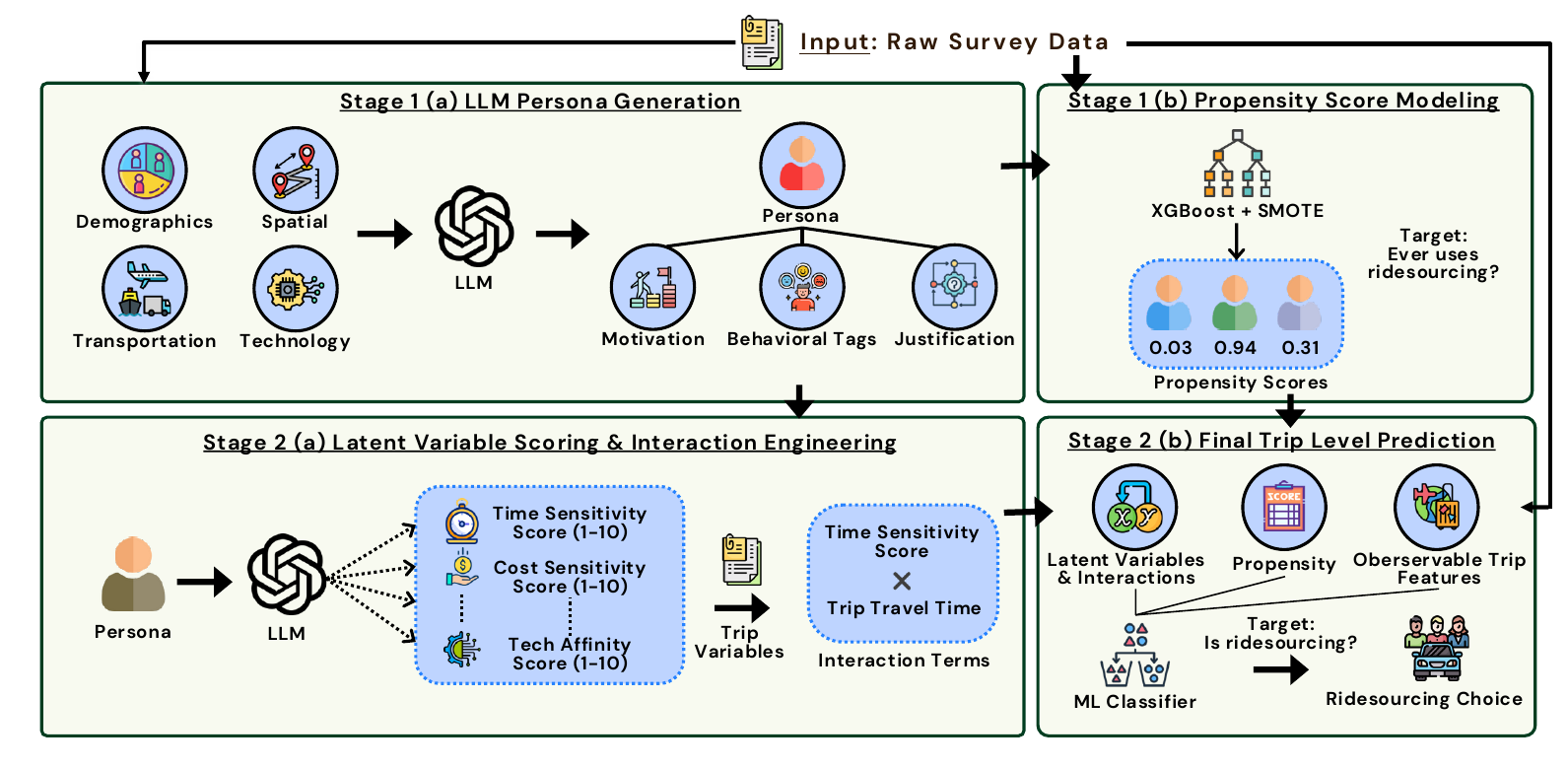}
    \caption{The two-stage hierarchical framework for ridesourcing choice prediction. Stage 1 models a traveler's individual-level propensity to use ridesourcing. Stage 2 then predicts the trip-level choice by integrating this propensity score with synthesized latent attitudes and situational trip features.}
    \label{fig:pipeline}
\end{figure*}

Our methodology, the \textit{Synthesizing Attitudes, Predicting Actions} (SAPA) framework, is a two-stage, hierarchical process designed to overcome the dual challenges of rare event prediction and missing psychological data. This framing is inspired by hierarchical behavioral models, which represent decision processes that occur in stages \citep{mcfadden1974}. Stage 1 models the higher-level, stable disposition of an individual toward ridesourcing at the \textbf{individual-level}. Stage 2 then models the lower-level, situational choice at the \textbf{trip-level}, conditional on the disposition synthesized in the first stage. A key contribution of this work is the rigorous validation of this framework on a large-scale travel survey. An overview of the data flow is presented in Figure \ref{fig:pipeline}.

\subsection{Stage 1: Individual-Level Propensity Score Modeling}
The first stage focuses on engineering a single, powerful feature that captures a traveler's general, long-term inclination toward using ridesourcing. This is accomplished by training a propensity score model on a \textbf{individual-level} target variable: \texttt{ever\_uses\_ridesourcing}. The modeling process integrates LLM-generated features to create a robust and generalizable score.

\subsubsection{LLM-Based Persona and Feature Generation}
For each of the 7,500 unique travelers across the four survey years in the PSRC dataset, we use a Large Language Model (Llama-3.1-8B-Instruct) prompted (see Appendix) to act as an expert cognitive behavioral analyst. The LLM synthesizes each traveler's observable demographic and household data into a structured JSON object containing a qualitative persona. This persona includes a concise primary motivation (e.g., "Time-sensitive professional") and a list of behavioral tags (e.g., "tech-savvy," "cost-conscious"). These tags are then processed into features for the propensity model.

\subsubsection{Propensity Score Modeling and the Impact of LLM Features}
The propensity score model is an XGBoost classifier \cite{chen2016xgboost} trained to predict \texttt{ever\_uses\_ridesourcing}, using the Synthetic Minority Over-sampling Technique (SMOTE) to handle class imbalance \citealp{chawla2002}. A critical innovation of this stage was to explicitly test the value of the LLM-inferred features for model generalization. A baseline model using only observable features overfit severely on the test set. Despite a nominal PR-AUC of 0.1463, it failed to identify a single true positive, resulting in 0.00 precision and recall. In stark contrast, enhancing the model with simple features derived from the LLM personas resolved this failure entirely. The enhanced model successfully identified 764 true positives, achieving a test precision of 0.3214 and recall of 0.4916. This resulted in a final PR-AUC of 0.3887—a \textbf{165.5\% improvement} over the baseline. This demonstrates that LLM-inferred signals are essential for building a generalizable model of an individual’s disposition. The final, calibrated probability from this superior model serves as the \texttt{propensity\_score} feature for Stage 2.

\subsection{Stage 2: Trip-Level Choice Prediction}
The second stage focuses on the trip-level ridesourcing mode choice prediction. This is accomplished by first using the LLM to generate additional behavioral features from the personas, and then integrating these features with the Stage 1 propensity score and observable trip data into a final machine learning classifier.

\subsubsection{Latent Variable Scoring and Interaction Engineering}
In a separate process, the unstructured text from the personas generated in Stage 1 is transformed into a structured, quantitative feature set. Using a second prompt (see Appendix), the LLM scores each traveler's persona on seven predefined latent variables. These variables, grounded in established behavioral theories, are defined in Table \ref{tab:latent_variables}.

To unlock the predictive value of these synthesized attitudes, they are not used as standalone main effects. Instead, six of the variables are multiplied by corresponding trip-level variables to create theory-driven interaction terms. These terms model how a stable, individual-level attitude is "activated" by a specific trip context. The operationalization of these interaction terms, informed by the feature engineering script, is also detailed in Table \ref{tab:latent_variables}.

\begin{table*}[h]
\centering
\small                              
\setlength{\tabcolsep}{3pt}         
\renewcommand{\arraystretch}{1.1}   
\begin{threeparttable}
\begin{tabular}{@{}p{0.20\linewidth}p{0.45\linewidth}ll@{}}
\toprule
\textbf{Latent Variable}    & \textbf{Definition}                                           & \textbf{Theory}    & \textbf{Interaction Term}        \\ 
\midrule
Time Sensitivity            & {\footnotesize A traveler’s prioritization of minimizing travel time.}        & TPB                & Score~×~traveltime        \\
Cost Sensitivity            & {\footnotesize A traveler’s prioritization of minimizing monetary cost.}      & TPB                & Score~×~fare             \\
Pro‐Car Attitude            & {\footnotesize A preference for private vehicles due to comfort or privacy.}  & TPB                & Score~×~household vehicles           \\
Convenience \& Comfort           & {\footnotesize A desire for seamless, low‐effort, and comfortable travel.}    & TPB                & Score~×~distance        \\
Environmental Concern       & {\footnotesize The degree to which a traveler considers environmental impacts.}& TPB               & Score~×~mandatory trip\tnote{a} \\
Spontaneity                 & {\footnotesize A tendency toward unplanned or impulsive travel decisions.}    & Exploratory        & Score~×~weekend indicator     \\
Technology Affinity         & {\footnotesize A traveler’s comfort and enthusiasm for using technology.}     & TAM                & (None)                           \\
\bottomrule
\end{tabular}
\begin{tablenotes}[flushleft]
  \item[a] \footnotesize{Indicates work or school trips.}
\end{tablenotes}
\caption{Definitions and Operationalization of Synthesized Latent Variables.}
\label{tab:latent_variables}
\end{threeparttable}
\end{table*}

These individual-level scores are merged with the trip-level dataset, assigning the same set of scores to every trip made by a given individual. The final feature set for the Stage 2 model comprises trip-level observables, the \texttt{propensity\_score} from Stage 1, the seven raw latent variable scores, and the six interaction terms.

\subsection{Model Specification}
The final trip-level choice model predicts the probability of ridesourcing ($y_{ni}=1$) for a trip $i$ by individual $n$. This probability is a function of trip-level observables ($\mathbf{x}_{ni}$), the individual-level propensity score ($s_n$), synthesized latent attitudes ($\mathbf{l}_n$), and their interactions:
\begin{equation}
  P(y_{ni}=1) = f_{\text{Classifier}}(\mathbf{x}_{ni}, s_n, \mathbf{l}_n, \mathbf{x}_{ni} \odot \mathbf{l}_n)
\end{equation}
where $f_{\text{Classifier}}$ is a machine learning classifier (e.g., LightGBM, CatBoost) and $\odot$ represents the element-wise creation of interaction terms. The ablation studies in the results section systematically evaluate the predictive contribution of each feature set across multiple classifiers.

\section{Experimental Setup}

\subsection{Data Sources and Processing}
We validate our framework's robustness on a large-scale household travel survey: the Puget Sound Regional Council's (PSRC) Household Travel Survey, conducted between 2017 and 2023 \citep{PSRC2017_2023}. This rich, multi-year dataset allows for a deep and rigorous test of our methodology.

The dataset underwent a multi-stage processing pipeline. This involved merging raw household, person, and trip files, followed by minimal data cleaning and comprehensive feature engineering. A critical step was the generation of spatial embeddings to capture the unique characteristics of each traveler's home location. We trained a Graph Attention Network (GAT) \cite{veličković2018graphattentionnetworks} on aggregated census tract-level data (e.g., population density, ridesourcing adoption) to learn a 128-dimension embedding, or ``spatial signature,'' for each tract. These embeddings were then merged back into the main trip dataset. Finally, the persona-level stratified splitting strategy described below was used to create the final modeling datasets, which are summarized in Table \ref{tab:data_stats}.

\begin{table}[h]
\centering                        
\setlength{\tabcolsep}{1mm}    
\renewcommand{\arraystretch}{1}
\begin{tabular}{@{}lcc@{}}
\toprule
\textbf{Statistic} & \textbf{PSRC} \\ 
\midrule
Period          & 2017–2023  \\
Train Trips     & 58,954     \\
Test Trips      & 15,052     \\
Ridesouring Trips (Train) (\%)    & 1.03\%    \\
Ridesourcing Trips (Test)  (\%)   & 1.26\%    \\
\bottomrule
\end{tabular}
\caption{Summary Statistics of the Final Processed Datasets for Trip‐Level Modeling.}
\label{tab:data_stats}
\end{table}

\begin{table*}[ht]
\centering
\begin{tabular}{@{}lcccc@{}}
\toprule
\textbf{Model Scenario} & \textbf{Best Model} & \textbf{PR-AUC} & \textbf{ROC-AUC} & \textbf{F1-Score} \\
\midrule
Baseline & RandomForest & 0.1409 & 0.7344 & 0.1976 \\
\textit{(Observables Only)} & & & & \\
\addlinespace
Best SAPA Model & LightGBM & \textbf{0.2479} & \textbf{0.8754} & \textbf{0.3000} \\
\textit{(All Features + All Interactions)} & & & & \\
\midrule
\textit{Relative Improvement (\%)} & & \textbf{+75.9\%} & \textbf{+19.2\%} & \textbf{+51.8\%} \\
\bottomrule
\end{tabular}
\caption{SAPA Framework Performance Summary on the PSRC Test Set. This table contrasts the best-performing baseline model (using only observable data) against the best-performing SAPA-enhanced model, highlighting key performance metrics.}
\label{tab:main_results}
\end{table*}

\subsubsection{Persona-Level Stratified Splitting}
To ensure a robust evaluation that tests generalization to unseen individuals and prevents data leakage, we performed a \textbf{persona-level stratified train-test split}. This strategy keeps all trips made by a single individual ("persona") together in either the training or testing set. The split was stratified based on whether a person had \textit{ever} used a ridesourcing service to maintain a similar distribution of the rare class across sets. For the multi-year PSRC dataset, stratification was further refined by the traveler's spatial profile and survey year to ensure temporal and geographic balance.

\subsection{Baseline Model and Evaluation Metrics}
Our primary baseline is the \textbf{strongest-performing state-of-the-art machine learning model trained using only observable features}. We tested several algorithms (RandomForest, XGBoost, LightGBM, CatBoost) to establish this robust performance benchmark. Given the severe class imbalance, our primary evaluation metric is the \textbf{Area Under the Precision-Recall Curve (PR-AUC)}, which is more informative than ROC-AUC for rare event prediction \citep{davis2006precisionrecall, fawcett2006roc, ke2017lightgbm, chen2016xgboost, prokhorenkova2018catboost, breiman2001randomforest}. We also report on \textbf{ROC-AUC}, \textbf{F1-Score}, and the \textbf{Matthews Correlation Coefficient (MCC)} to provide a holistic view of model performance.

\section{Results and Analysis}

Our comprehensive ablation study reveals the SAPA framework's significant predictive power and robustness, demonstrating substantial improvements over strong baselines while uncovering important, context-specific drivers of ridesourcing choice. The findings presented here are based on the held-out test set. However, they were validated by a rigorous 5-fold cross-validation that confirmed the same performance patterns, as detailed in the Appendix.

\subsection{Overall Performance and Framework Validation}
The SAPA framework delivers a significant performance lift over strong baseline models. As summarized in Table \ref{tab:main_results}, the framework's ability to enrich the dataset with synthesized psychological features leads to markedly better predictions of the rare event of choosing ridesourcing.

For the PSRC dataset, the improvements are substantial across all key metrics. The best SAPA model, using a LightGBM classifier, achieves a test set PR-AUC of 0.2479, representing a prominent \textbf{75.9\%} improvement over the strongest baseline. This is accompanied by a \textbf{51.8\%} increase in F1-Score and a \textbf{19.2\%} increase in ROC-AUC, demonstrating a comprehensive performance gain.

\subsection{Cross-Sectional Validation of Synthesized Latent Variables}

\begin{figure}[t]
    \centering
    \includegraphics[width=.75\columnwidth]{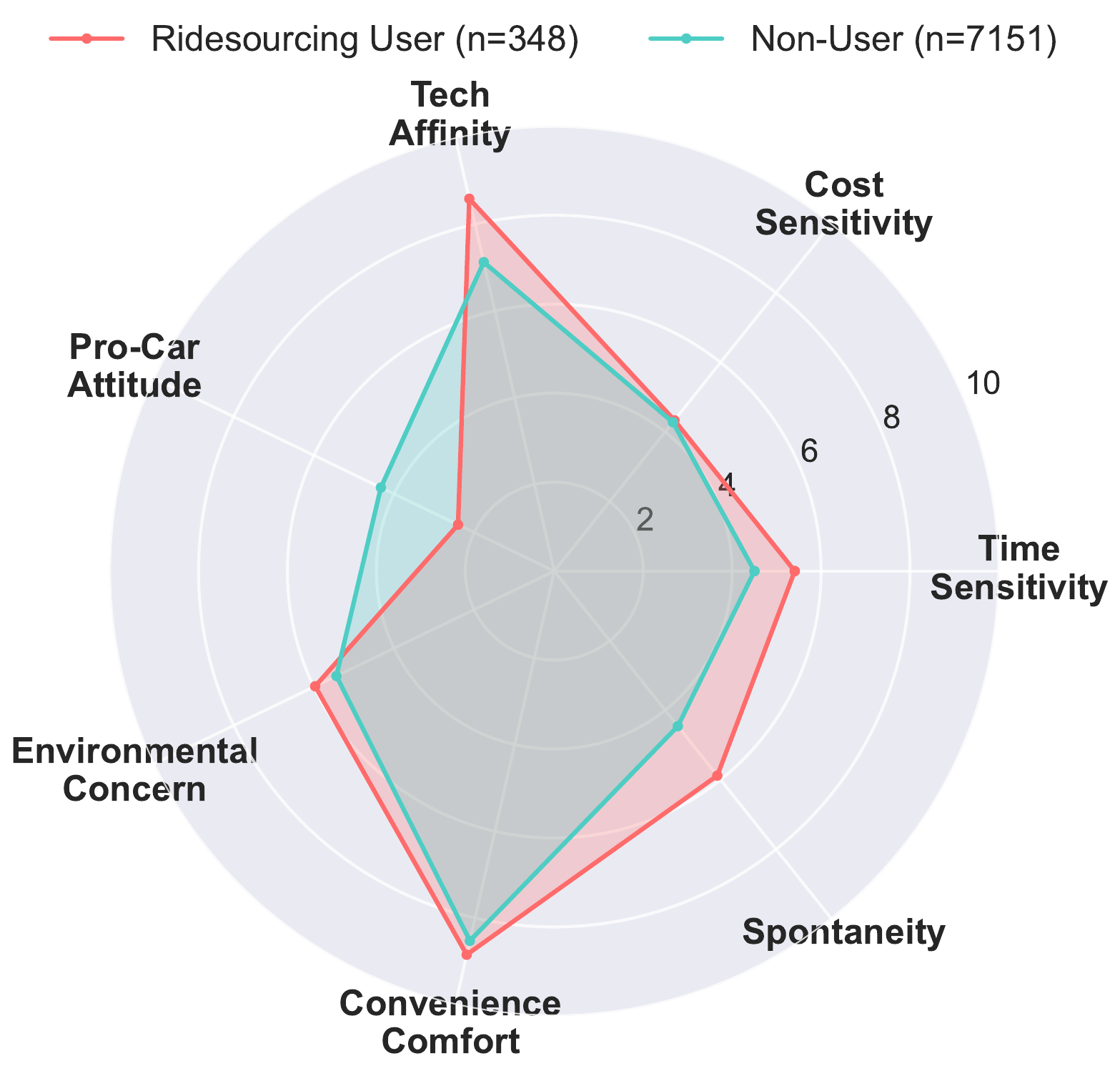}
    \caption{A comparison of mean latent variable profiles for ridesourcing users (n=348) and non-users (n=7151) in the PSRC dataset.}
    \label{fig:latent_var_validation}
\end{figure}

The key to the SAPA framework is the generated personas (Stage 1 shown in Figure \ref{fig:pipeline}). Therefore, an important step of our framework is to investigate the validity of the generated personas, e.g., if the personas are aligned with existing ridesourcing choice-making behavior. We use cross-sectional analysis to understand how latent variables of the persona vary across different traveler segments (e.g., ridesourcing user vs. non-user, ridesourcing users in different age or income segments). Figure \ref{fig:latent_var_validation} presents a radar chart comparing the personas across ridesourcing users and non-users along seven latent dimensions. Each axis of the chart represents a score for a specific latent trait. The results indicate that travelers who are more likely to use ridesourcing services tend to score higher in tech affinity, time sensitivity, spontaneity, environmental concern, and convenience/comfort, while scoring lower in pro-car attitude. This pattern is consistent with prior literature on ridesourcing adoption behavior and supports the validity of the generated personas \cite{zhang2022machine,marquet2020spatial,yu2019exploring,zhang2024analyzing,ghaffar2020modeling}. In the Appendix, we also present an additional comparison of how the latent variable score of ridesourcing users vary across age, income and vehicle availability segments. Results reveal that younger users ($<35$) exhibit higher tech affinity and spontaneity \cite{wang2019ridesourcing}, while higher-income users tend to be less cost-sensitive but more comfort- and tech-oriented \cite{soria2020k,zhang2024analyzing,yan2020using}. Additionally, users without a personal vehicle show stronger environmental concern and cost sensitivity \cite{asgari2020propensity,azimi2020role}. All these findings demonstrate that the generated personas can capture key behavioral traits in ridesourcing choice-making process. 

\subsection{Robustness Across Classifiers}
A key finding of our expanded analysis is that the SAPA framework is not dependent on a single classifier. Rather, its synthesized features provide a consistent and substantial performance boost to a wide range of state-of-the-art algorithms. Table \ref{tab:robustness_psrc} shows the impact of adding the full suite of SAPA features to four different classifiers on the PSRC test set.

In every case, adding the SAPA features resulted in a significant improvement across all metrics. For instance, LightGBM's PR-AUC increased by 84.2\%, while XGBoost's F1-Score nearly doubled with a 94.5\% increase. This demonstrates that the LLM-synthesized features capture true behavioral signals that are valuable to any underlying predictive model, confirming the robustness and general utility of our framework.

Wilcoxon signed-rank tests \cite{Woolson2008_Wilcoxon} confirm that the SAPA framework significantly improves PR-AUC performance over the baseline for all four classifiers (p = 0.0312 for all models), with effect sizes (Cohen's d) ranging from \textbf{2.80 to 3.29}, indicating very large practical significance.

\begin{table}[h]
\centering
\begin{threeparttable}
\small
\setlength{\tabcolsep}{8pt} 
\begin{tabular}{@{}lccc@{}}
\toprule
\textbf{Classifier + Features} & \textbf{PR-AUC} & \textbf{F1-Score} & \textbf{ROC-AUC} \\
\midrule

\textbf{RandomForest} &  &  &  \\
Observables Only & 0.1409 & 0.1976 & 0.7344 \\
+ SAPA Features & \textbf{0.2445} & \textbf{0.2929} & \textbf{0.8357} \\
\textit{Improvement (\%)} & \textit{+73.5\%} & \textit{+48.2\%} & \textit{+13.8\%} \\

\addlinespace
\textbf{XGBoost} &  &  &  \\
Observables Only & 0.1394 & 0.1520 & 0.8411 \\
+ SAPA Features & \textbf{0.2092} & \textbf{0.2957} & \textbf{0.8797} \\
\textit{Improvement (\%)} & \textit{+49.9\%} & \textit{+94.5\%} & \textit{+4.6\%} \\

\addlinespace
\textbf{LightGBM} &  &  &  \\
Observables Only & 0.1346 & 0.1880 & 0.8211 \\
+ SAPA Features & \textbf{0.2479} & \textbf{0.3000} & \textbf{0.8754} \\
\textit{Improvement (\%)} & \textit{+84.2\%} & \textit{+59.6\%} & \textit{+6.6\%} \\

\addlinespace
\textbf{CatBoost} &  &  &  \\
Observables Only & 0.1361 & 0.1619 & 0.8356 \\
+ SAPA Features & \textbf{0.2037} & \textbf{0.2864} & \textbf{0.8808} \\
\textit{Improvement (\%)} & \textit{+49.7\%} & \textit{+76.9\%} & \textit{+5.4\%} \\

\bottomrule
\end{tabular}
\begin{tablenotes}[para,flushleft]
  \footnotesize{“+ SAPA Features” corresponds to \textit{All Features + All Interactions}.}
\end{tablenotes}
\caption{SAPA Framework Performance Boost Across Different Classifiers (PSRC Test Set)}
\label{tab:robustness_psrc}
\end{threeparttable}
\end{table}

\subsection{Component Analysis and Context-Dependent Complexity}
An ablation study on the PSRC dataset using the best-performing classifier (LightGBM) reveals the contribution of each component of the SAPA framework (Table \ref{tab:psrc_ablation}). Adding only the latent variable scores provides a marginal improvement over the baseline, but the Stage 1 propensity score is a powerful feature, significantly boosting PR-AUC from 0.1346 to 0.2060 when added to the observables. The full SAPA framework, combining all components, yields the best results, with the most complex model including all interaction terms achieving the highest PR-AUC (0.2479) and F1-Score (0.3000). This demonstrates a clear and consistent performance benefit from modeling not just the synthesized attitudes, but also the theory-driven interactions that capture how those attitudes are activated by specific trip characteristics.

The results highlight a key finding: for a rich, high-quality dataset like PSRC, increasing model complexity by incorporating nuanced behavioral features leads to superior predictive performance. The success of the most complex model suggests that capturing the interplay between stable personality traits and situational factors is crucial for accurately predicting rare choices.

\begin{table}[ht]
\centering
\begin{threeparttable}
\small
\setlength{\tabcolsep}{5pt}
\begin{tabular}{@{}lccc@{}}
\toprule
\textbf{Model Components} & \textbf{PR-AUC} & \textbf{F1-Score} & \textbf{MCC} \\
\midrule
\textbf{1. Observables Only} & 0.1346 & 0.1880 & 0.2010 \\
\textit{(Baseline)} & & & \\
\addlinespace
\textbf{2. + Latent Only} & 0.1464 & 0.1849 & 0.1861 \\
\textit{(Observables + Latent)} & & & \\
\addlinespace
\textbf{3. + Propensity Only} & 0.2060 & 0.2722 & 0.2638 \\
\textit{(Observables + Propensity)} & & & \\
\addlinespace
\textbf{4. Full SAPA} & 0.2167 & 0.2994 & \textbf{0.2987} \\
\textit{(No Interactions)} & & & \\
\addlinespace
\textbf{5. Full SAPA} & \textbf{0.2479} & \textbf{0.3000} & 0.2911 \\
\textit{(All Interactions)} & & & \\
\bottomrule
\end{tabular}
\caption{Ablation Study of SAPA Components on the PSRC Dataset (LightGBM Test Set).}
\label{tab:psrc_ablation}
\end{threeparttable}
\end{table}

\section{Discussion}
Our results strongly support the SAPA framework and offer several key insights for the future of behavioral modeling in data-scarce, rare-event contexts.

\subsection{A Validated Solution to the Latent Variable Problem}
Our results validate that LLMs can synthesize latent variables with measurable predictive power. As demonstrated in the Stage 1 propensity model, the synthesized features were essential for solving a critical overfitting problem. The subsequent success of the full trip-level model confirms that these attitudes capture meaningful behavioral signals, an observation supported by the framework's consistent performance boost across multiple advanced classifiers (Table \ref{tab:robustness_psrc}). This underscores the intrinsic value of the synthesized features themselves, independent of the final classifier.

\subsection{Modeling Behavioral Complexity: The Role of Latent Interactions}
Our deep validation on the PSRC dataset reveals that the influence of synthesized attitudes is complex and interactive. The ablation study (Table \ref{tab:psrc_ablation}) shows that performance continuously improves as more sophisticated behavioral features are added, with the best model being the one that utilizes all synthesized features and all of their interaction terms. This finding strongly suggests that for accurately modeling behavior, it is not enough to know a person's general attitudes (e.g., being time-sensitive), it is critical to model how those attitudes are activated by the context of a specific choice (e.g., how time-sensitivity interacts with the actual travel time of a trip). This underscores the importance of theory-driven feature engineering and highlights the framework's utility for uncovering nuanced behavioral drivers and provides a clear methodology for moving beyond static psychometric profiles to dynamic, context-aware behavioral models.

\subsection{A New Methodological Blueprint}
This work offers a new methodological blueprint for computational social science. By grounding LLM-based synthesis in established theory and validating the outputs through predictive performance on a suite of models, the SAPA framework provides a robust template for overcoming data scarcity. This approach can be extended to other domains where latent, unobserved factors are critical drivers of behavior but are absent from large-scale observational data, such as predicting technology adoption, public health choices, or consumer behavior.

\section{Ethical Considerations and Broader Impact}
The use of LLMs to infer personal attitudes raises important ethical considerations. While the models are trained on anonymized survey data, the generation of psychological profiles requires careful handling to ensure privacy. More importantly, LLMs can inherit and amplify societal biases present in their training data \citep{veselovsky2023, veselovsky2024synthetic}. An LLM might generate stereotypical personas based on demographic attributes, which could lead to discriminatory model outcomes. Future work must implement formal bias auditing and mitigation techniques to address these risks.

\section{Limitations and Future Work}
This study has several limitations. First, the methodology's reliance on LLM-generated personas introduces sensitivity to the chosen LLM and prompt engineering. Second, the LLM-generated latent variable scores are \textit{synthetic approximations}, not ground-truth psychological measurements. Their validity requires further research, ideally through comparison with data from traditional psychometric surveys. Finally, while our deep validation on a single, rich dataset is a key strength, a critical next step is to test the framework's transferability to other urban contexts. Future work should explore the interplay between data quality, feature engineering, and the choice of the final machine learning classifier across different geographic and cultural settings.

\section{Conclusion}
This paper demonstrates that Large Language Models can successfully overcome the long-standing problem of missing psychological data in transportation choice modeling. We introduced and validated the SAPA framework, a novel hierarchical method that synthesizes latent attitudes from standard survey data to significantly improve the prediction of rare travel choices. Our comprehensive ablation studies, conducted on a large-scale, multi-year travel survey, show the full SAPA framework significantly outperforms strong baselines—improving PR-AUC by up to 75.9\%—with its LLM-synthesized features proving essential for model generalization and robustly enhancing performance across multiple state-of-the-art classifiers. Furthermore, the framework provides deeper behavioral insights, revealing the critical importance of modeling the complex, interactive relationship between latent attitudes and situational factors. By offering a practical and reproducible blueprint, this work paves the way for integrating sophisticated, AI-driven behavioral insights into choice models, opening new avenues for both computational social science and more effective public policy.

\section{Acknowledgements}
This material is based upon work supported by the U.S. National Science Foundation (NSF) under award No. 2338959. Any opinions, findings, conclusions, or recommendations expressed in this material are those of the authors and do not necessarily reflect the views of NSF. During the preparation of this work, the authors used ChatGPT in order to check grammar errors and improve the language. After using this tool, the authors reviewed and edited the content as needed and took full responsibility for the content of the publication.

\setcounter{secnumdepth}{1}
\begin{appendix}

\newpage%
\renewcommand{\thesection}{\Alph{section}}

\section{Dataset and Pre-processing}

\subsection{Dataset Description}
The experiments in this study are based on the Puget Sound Regional Council (PSRC) Household Travel Survey data, collected between 2017 and 2023. This survey is a comprehensive regional effort to understand travel behavior and patterns in the Seattle metropolitan area. The dataset is publicly available and comprises several linked tables. Our analysis primarily utilizes the four tables described in Table \ref{tab:dataset_sources}.

\begin{table}[h!]
\centering
\begin{tabular}{@{}lp{0.6\columnwidth}@{}}
\toprule
\textbf{Data File} & \textbf{Description and Key Variables Used} \\
\midrule
Households & Contains one record per household. Used for household-level features such as \texttt{hhincome\_detailed}, \texttt{vehicle\_count}, and home location. \\
\addlinespace
Persons & Contains one record for each person within a surveyed household. Used for person-level features like \texttt{age}, \texttt{gender}, \texttt{employment}, and \texttt{education}. \\
\addlinespace
Vehicles & Contains records for each vehicle owned by a household. Used primarily to confirm \texttt{vehicle\_count}. \\
\addlinespace
Trips & The core travel diary data, containing one record per trip made by a person. Used for trip-level features like \texttt{travel\_mode}, \texttt{origin/destination}, and \texttt{trip\_purpose}. \\
Days & Contains day-level survey information and travel diary metadata. Used for temporal context and survey validation. \\
\bottomrule
\end{tabular}
\caption{PSRC Household Travel Survey Data Sources.}
\label{tab:dataset_sources}
\end{table}

\subsection{Data Pre-processing and Feature Engineering}
The raw PSRC data was processed through a multi-stage pipeline to prepare it for modeling.
\begin{enumerate}
    \item \textbf{Data Merging:} The raw household, person, and trip files were merged into a single trip-level dataset, with each row representing a unique trip and containing all associated person and household attributes.
    \item \textbf{Data Cleaning:} Missing values were handled using standard imputation techniques or by treating them as a separate category where appropriate. Inconsistent records were removed.
    \item \textbf{Feature Engineering:} A wide range of features were engineered from the raw data. This included creating categorical variables (e.g., `age\_group` from `age`), calculating ratios (e.g., `vehicles\_per\_person`), and one-hot encoding categorical features for machine learning models.
    \item \textbf{Spatial Embeddings:} To capture geographic context, a Graph Attention Network (GAT) was trained on census tract-level data to generate a 128-dimension "spatial signature" for each traveler's home location. These embeddings were then merged into the final feature set.
\end{enumerate}

\section{LLM Framework Details}

\subsection{Prompt for Persona Generation (Stage 1)}
A system prompt (Listing \ref{lst:persona_system_prompt}) and a user prompt (Listing \ref{lst:persona_user_prompt}) were combined for the persona generation task.

\begin{listing}[h!]
\begin{lstlisting}[language=]
You are an expert cognitive behavioral analyst. Your task is to analyze a traveler's profile to infer their likely travel behavior and motivations. Focus on clues that might reveal underlying attitudes related to time, cost, technology, and transportation preferences.

You must provide your output *only* as a single, strictly valid JSON object with three specific keys: 'primary_motivation', 'behavioral_tags', and 'brief_justification'. Do not add any conversational text or markdown formatting around the JSON object.
\end{lstlisting}
\caption{System Prompt for Persona Generation}
\label{lst:persona_system_prompt}
\end{listing}

\begin{listing}[h!]
\begin{lstlisting}[language=]
Analyze the provided traveler data. Based on this data, distill the agent's persona into its core components and provide the output as a single-line, strictly valid JSON object.

**INPUT DATA:**
* **Socio-Demographics:** {socio_demographics}
* **Home Environment Profile:** {spatial_profile_label}
* **Spatial Context Vector:** {spatial_embeddings_json}

**YOUR TASK:**
Return a single JSON object with the following structure:
{{
  "primary_motivation": "[A concise phrase (max 5 words) describing the main driver for this person's travel choices]",
  "behavioral_tags": ["[A list of 3-5 relevant keyword tags]", "[tag2]", "..."],
  "brief_justification": "[A single sentence (max 25 words) explaining your reasoning]"
}}
\end{lstlisting}
\caption{User Prompt for Persona Generation}
\label{lst:persona_user_prompt}
\end{listing}

\subsection{Prompt for Latent Variable Scoring (Stage 2)}
The user prompt as shown in Listing \ref{lst:latent_prompt} was used to score the personas generated in Stage 1.

\begin{listing*}[h!]
\begin{lstlisting}[language=]
**TASK:** Read the following traveler profile. For each of the seven behavioral dimensions, assign a score from 1 to 10 based on the provided scoring guide.

**SCORING GUIDE:**
- **Time Sensitivity:** 1 = Highly patient, willing to wait for cheaper options. 5 = Balances time and cost. 10 = Prioritizes speed above all else, willing to pay for it.
- **Cost Sensitivity:** 1 = Spends freely for convenience/speed. 5 = Balances cost and other factors. 10 = Exclusively seeks the cheapest option.
- **Tech Affinity:** 1 = Avoids technology and apps. 5 = Uses common apps but not an early adopter. 10 = Eagerly adopts new technology and app-based services.
- **Pro Car Attitude:** 1 = Prefers any mode over driving. 5 = Views car as one of many options. 10 = Exclusively prefers the privacy and control of a personal car.
- **Environmental Concern:** 1 = Unconcerned with environmental impact. 5 = Aware of impact but prioritizes other factors. 10 = Actively chooses sustainable/eco-friendly options.
- **Convenience/Comfort Seeking:** 1 = Tolerates inconvenience for cost/speed. 5 = Appreciates convenience but makes trade-offs. 10 = Main motivation is the easiest, most comfortable travel.
- **Spontaneity:** 1 = Plans all trips far in advance. 5 = Plans most trips but can be spontaneous. 10 = Makes unplanned, last-minute travel decisions.

---
**EXAMPLE:** ... (example omitted for brevity)
---

**YOUR TASK:**
**INPUT PROFILE:**
- Primary Motivation: {motivation}
- Behavioral Tags: {tags}
- Brief Justification: {justification}
**OUTPUT:** (JSON object with scores and justifications for all 7 dimensions)
\end{lstlisting}
\caption{User Prompt for Latent Variable Scoring}
\label{lst:latent_prompt}
\end{listing*}

\subsection{Qualitative Examples of Generated Personas}
\label{sec:qualitative_examples}

To illustrate the richness and diversity of the LLM-generated personas, we present three anonymized examples from Stage 1 of the SAPA framework. These examples demonstrate how the LLM synthesizes socio-demographic, spatial, and behavioral data into coherent travel personas.

\subsubsection{Example 1: Tech-Savvy Urban Professional}
This persona (Listing \ref{lst:tech_savvy_persona}) was generated for a middle-aged individual residing in a medium income household with few vehicles. The LLM correctly identified the individual's high time sensitivity (9/10) and low cost sensitivity (2/10), while noting their moderate environmental concern (4/10) and high spontaneity (8/10), reflecting their flexible travel preferences.

\begin{listing}[h!]
\caption{Tech-Savvy Urban Professional Persona}
\label{lst:tech_savvy_persona}
\begin{lstlisting}[language={}]
{
  "primary_motivation": "Convenience and flexibility in travel",
  "behavioral_tags": ["Tech-savvy", "Urban", "Affluent", "Time-efficient", "Flexible"],
  "brief_justification": "This individual prioritizes convenience and flexibility in their travel choices, leveraging technology and affluence to optimize their time and experiences."
}
\end{lstlisting}
\end{listing}

\subsubsection{Example 2: Environmentally Conscious Urban Explorer}
This example demonstrates the LLM's ability to identify environmental consciousness from behavioral patterns. This middle-aged individual (Listing \ref{lst:environmentally_conscious_persona}), residing in a low income household, was characterized by their preference for walking and public transportation, leading to maximum environmental concern scores (10/10) and moderate cost sensitivity (4/10), with high spontaneity (8/10) reflecting their exploratory nature.

\begin{listing}[h!]
\caption{Environmentally Conscious Urban Explorer Persona}
\label{lst:environmentally_conscious_persona}
\begin{lstlisting}[language={}]
{
  "primary_motivation": "Environmentally conscious urban explorer",
  "behavioral_tags": ["sustainability", "walkability", "urban exploration", "smartphone user", "middle-aged"],
  "brief_justification": "This persona prioritizes environmentally friendly transportation options, such as walking, and is likely to explore urban areas using their smartphone, driven by a desire for sustainability and urban exploration."
}
\end{lstlisting}
\end{listing}

\subsubsection{Example 3: Comfort and Convenience Driven Senior}
This senior individual (Listing \ref{lst:comfort_convenience_persona}), residing in a low income household with few vehicles, exemplifies how the LLM captures age-related travel preferences. The LLM identified their car-dependent lifestyle and preference for comfort, resulting in low time sensitivity (3/10) and moderate spontaneity (5/10), with low cost sensitivity (2/10) reflecting their affluent status despite being in a lower income group.

\begin{listing}[h!]
\caption{Comfort and Convenience Driven Senior Persona}
\label{lst:comfort_convenience_persona}
\begin{lstlisting}[language={}]
{
  "primary_motivation": "Comfort and convenience driven",
  "behavioral_tags": ["affluent", "car-dependent", "tech-savvy", "urban", "Seattle"],
  "brief_justification": "This individual's high income, graduate education, and smartphone ownership suggest a preference for comfort and convenience in their travel choices, likely prioritizing driving and using urban amenities."
}
\end{lstlisting}
\end{listing}

These examples illustrate the framework's ability to generate nuanced, contextually appropriate personas that capture both explicit behavioral patterns and implicit psychological traits, providing a rich foundation for the subsequent latent variable scoring and propensity modeling stages.

\section{Computational Details and Reproducibility}

\subsection{Model Hyperparameters}
For the ablation studies, we used fixed hyperparameter configurations for all models rather than conducting systematic hyperparameter optimization. This approach was chosen to maintain consistency across ablation scenarios and focus on the comparative contribution of different feature sets. The parameters, listed in Table \ref{tab:hyperparameters}, were set to commonly recommended default or baseline values for each algorithm.

\begin{table}[h!]
\centering
\begin{tabular}{@{}ll@{}}
\toprule
\textbf{Model} & \textbf{Hyperparameters} \\
\midrule
\multirow{3}{*}{RandomForest} & `n\_estimators=100` \\
& `class\_weight='balanced'` \\
& `random\_state=42` \\
\addlinespace
\multirow{5}{*}{XGBoost} & `n\_estimators=100`, `max\_depth=6` \\
& `learning\_rate=0.1` \\
& `subsample=0.8`, `colsample\_bytree=0.8` \\
& `scale\_pos\_weight` \\
& (dynamic: neg\_count/pos\_count) \\
& `random\_state=42` \\
\addlinespace
\multirow{6}{*}{LightGBM} & `n\_estimators=100`, `max\_depth=6` \\
& `learning\_rate=0.1` \\
& `subsample=0.8`, `colsample\_bytree=0.8` \\
& `scale\_pos\_weight` \\
& (dynamic: neg\_count/pos\_count) \\
& `random\_state=42`, `verbose=-1` \\
\addlinespace
\multirow{5}{*}{CatBoost} & `iterations=100`, `depth=6` \\
& `learning\_rate=0.1` \\
& `loss\_function='Logloss'` \\
& `class\_weights` \\
& (dynamic: \{0: 1.0, 1: neg\_count/pos\_count\}) \\
& `random\_seed=42`, `verbose=False` \\
\bottomrule
\end{tabular}
\caption{Fixed Hyperparameters for All Models.}
\label{tab:hyperparameters}
\end{table}

\subsection{Class Balancing Strategy}
All models use dynamic class balancing to handle the imbalanced nature of ridesourcing prediction (where positive cases are rare). For tree-based models (XGBoost, LightGBM), this is achieved through `scale\_pos\_weight` calculated as the ratio of negative to positive samples. For CatBoost, this is implemented through `class\_weights` dictionary. RandomForest uses `class\_weight='balanced'` which automatically adjusts class weights inversely proportional to class frequencies.

\subsection{LLM and Computing Infrastructure}
Experiments were conducted on a workstation with an Apple M4 Pro CPU and 24GB of RAM. Key software versions include Python 3.10, scikit-learn 1.3+, LightGBM 4.1+, and XGBoost 2.0+. LLM API calls were made to GPT-3.5-turbo via the OpenAI API, with temperature settings configured for consistency (0.7 for creative persona generation, 0.3 for deterministic latent scoring).

\section{Evaluation Metrics}
The following metrics were used to evaluate model performance, chosen for their relevance to imbalanced classification problems. Let TP, TN, FP, and FN be the number of true positives, true negatives, false positives, and false negatives, respectively.

\begin{itemize}
    \item \textbf{Precision:} The fraction of positive predictions that are correct.
    \begin{equation}
        \text{Precision} = \frac{\text{TP}}{\text{TP} + \text{FP}}
    \end{equation}
    \item \textbf{Recall (Sensitivity):} The fraction of actual positives that are correctly identified.
    \begin{equation}
        \text{Recall} = \frac{\text{TP}}{\text{TP} + \text{FN}}
    \end{equation}
    \item \textbf{F1-Score:} The harmonic mean of precision and recall.
    \begin{equation}
        \text{F1-Score} = 2 \cdot \frac{\text{Precision} \cdot \text{Recall}}{\text{Precision} + \text{Recall}}
    \end{equation}
    \item \textbf{PR-AUC:} The Area Under the Precision-Recall Curve. This is the primary metric, as it provides a comprehensive summary of performance across all decision thresholds for imbalanced datasets.
    \item \textbf{MCC:} The Matthews Correlation Coefficient is a robust metric for binary classification that is less sensitive to class imbalance. It returns a value between -1 and +1.
    \begin{equation}
        \text{MCC} = \frac{\text{TP} \cdot \text{TN} - \text{FP} \cdot \text{FN}}{\sqrt{(\text{TP}+\text{FP})(\text{TP}+\text{FN})(\text{TN}+\text{FP})(\text{TN}+\text{FN})}}
    \end{equation}
\end{itemize}

\section{Detailed Experimental Results}

\subsection{Ablation Study Results}
This section provides the detailed results of the ablation study across all four primary classifiers. Table \ref{tab:ablation_cv_all_models} presents the 5-fold cross-validation results, and Table \ref{tab:ablation_test_all_models} presents the results on the held-out test set.

\begin{table*}[h!]
\centering
{%
\begin{tabular}{@{}lcccc@{}}
\toprule
\textbf{Model Components} & \textbf{RandomForest} & \textbf{XGBoost} & \textbf{LightGBM} & \textbf{CatBoost} \\
\midrule
1. Observables Only (Baseline) & 0.212 / 0.263 & 0.192 / 0.216 & 0.177 / 0.218 & 0.183 / 0.219 \\
\addlinespace
2. + Latent Only & 0.341 / 0.379 & 0.225 / 0.259 & 0.228 / 0.275 & 0.216 / 0.258 \\
\addlinespace
3. + Propensity Only & 0.350 / 0.390 & 0.291 / 0.340 & 0.293 / 0.333 & 0.280 / 0.326 \\
\addlinespace
4. Full SAPA (No Interactions) & 0.407 / 0.439 & 0.314 / 0.370 & 0.308 / 0.349 & 0.287 / 0.329 \\
\addlinespace
5. Full SAPA (All Interactions) & \textbf{0.419 / 0.438} & \textbf{0.307 / 0.355} & \textbf{0.300 / 0.343} & \textbf{0.292 / 0.334} \\
\bottomrule
\end{tabular}%
}
\caption{Ablation Study: 5-Fold Cross-Validation Results Across All Classifiers (PR-AUC / F1-Score).}
\label{tab:ablation_cv_all_models}
\end{table*}

\begin{table*}[h!]
\centering
{%
\begin{tabular}{@{}lcccc@{}}
\toprule
\textbf{Model Components} & \textbf{RandomForest} & \textbf{XGBoost} & \textbf{LightGBM} & \textbf{CatBoost} \\
\midrule
1. Observables Only (Baseline) & 0.141 / 0.198 & 0.139 / 0.152 & 0.135 / 0.188 & 0.136 / 0.162 \\
\addlinespace
2. + Latent Only & 0.164 / 0.207 & 0.165 / 0.186 & 0.146 / 0.185 & 0.122 / 0.169 \\
\addlinespace
3. + Propensity Only & 0.177 / 0.248 & 0.193 / 0.288 & 0.206 / 0.272 & 0.144 / 0.229 \\
\addlinespace
4. Full SAPA (No Interactions) & 0.233 / 0.288 & 0.217 / 0.294 & 0.217 / 0.299 & 0.189 / 0.278 \\
\addlinespace
5. Full SAPA (All Interactions) & \textbf{0.245 / 0.293} & \textbf{0.209 / 0.296} & \textbf{0.248 / 0.300} & \textbf{0.204 / 0.286} \\
\bottomrule
\end{tabular}%
}
\caption{Ablation Study: Held-Out Test Set Results Across All Classifiers (PR-AUC / F1-Score).}
\label{tab:ablation_test_all_models}
\end{table*}

\subsection{Statistical Significance Testing}
\textbf{Note on Experimental Re-run:} The cross-validation results presented in this section were generated from a dedicated experimental run designed to capture the performance score from each of the 5 folds for statistical testing. Minor variations from the main ablation results are expected due to stochasticity.

To formally validate performance gains, one-tailed Wilcoxon signed-rank tests were performed on the 5-fold CV results, comparing the baseline against the full SAPA model. As shown in Table \ref{tab:wilcoxon_results}, the SAPA framework provided a statistically significant and practically meaningful improvement for all models.

\begin{table*}[h!]
\centering

\begin{tabular}{@{}lcccc@{}}
\toprule
\textbf{Classifier} & \textbf{Baseline PR-AUC (Mean)} & \textbf{SAPA PR-AUC (Mean)} & \textbf{p-value} & \textbf{Effect Size (d)} \\
\midrule
RandomForest & 0.2132 & 0.4188 & 0.0312 & 3.197 \\
XGBoost & 0.1947 & 0.3030 & 0.0312 & 2.795 \\
LightGBM & 0.1803 & 0.3067 & 0.0312 & 3.279 \\
CatBoost & 0.1938 & 0.2817 & 0.0312 & 3.287 \\
\bottomrule
\end{tabular}
\caption{Wilcoxon Signed-Rank Test Results (Baseline vs. Full SAPA Model).}
\label{tab:wilcoxon_results}
\end{table*}

\subsubsection{Individual Fold Scores}
Table \ref{tab:individual_fold_scores} provides the detailed fold-by-fold PR-AUC scores that were used for the statistical testing, demonstrating consistent improvement in every fold for every classifier.

\begin{table}[h!]
\centering
\begin{tabular}{@{}lccccc@{}}
\toprule
\textbf{Model} & \textbf{Fold 1} & \textbf{Fold 2} & \textbf{Fold 3} & \textbf{Fold 4} & \textbf{Fold 5} \\
\midrule
\multicolumn{6}{l}{\textbf{RandomForest}} \\
\quad Baseline & 0.214 & 0.211 & 0.226 & 0.251 & 0.164 \\
\quad SAPA & 0.363 & 0.339 & 0.438 & 0.556 & 0.398 \\
\midrule
\multicolumn{6}{l}{\textbf{XGBoost}} \\
\quad Baseline & 0.180 & 0.185 & 0.201 & 0.232 & 0.175 \\
\quad SAPA & 0.307 & 0.230 & 0.338 & 0.357 & 0.282 \\
\midrule
\multicolumn{6}{l}{\textbf{LightGBM}} \\
\quad Baseline & 0.139 & 0.168 & 0.191 & 0.238 & 0.165 \\
\quad SAPA & 0.303 & 0.252 & 0.314 & 0.363 & 0.303 \\
\midrule
\multicolumn{6}{l}{\textbf{CatBoost}} \\
\quad Baseline & 0.199 & 0.180 & 0.193 & 0.230 & 0.168 \\
\quad SAPA & 0.310 & 0.268 & 0.296 & 0.298 & 0.236 \\
\bottomrule
\end{tabular}
\caption{Individual 5-Fold Cross-Validation PR-AUC Scores Used for Statistical Testing.}
\label{tab:individual_fold_scores}
\end{table}

\subsection{Feature Importance and Runtime Analysis}
\begin{figure*}[h!]
\centering
\includegraphics[width=1.5\columnwidth]{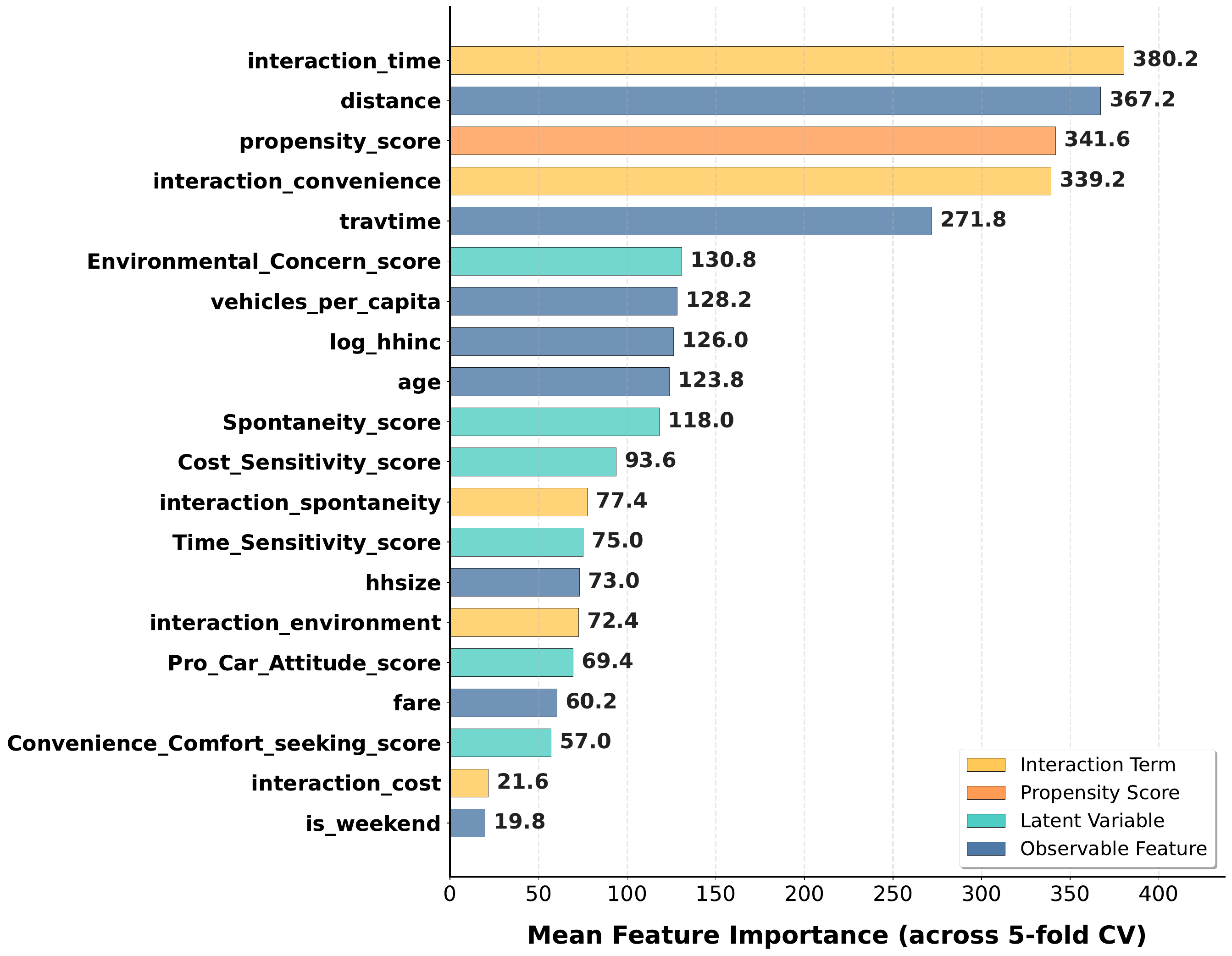}
\caption{Feature importance analysis for the LightGBM model showing the relative contribution of different feature categories in the full SAPA framework.}
\label{fig:feature_importance}
\end{figure*}

Figure \ref{fig:feature_importance} shows the relative importance of different feature categories for the LightGBM model. The analysis reveals that the LLM-generated features (Interaction Terms, Propensity Score, and Latent Variables) collectively contribute \textbf{60.3\%} of the total feature importance, underscoring their predictive value. This contribution is broken down into \textbf{Interaction Terms (30.2\%)}, \textbf{Latent Variables (18.5\%)}, and the \textbf{Propensity Score (11.6\%)}, with traditional observable features accounting for the remaining 39.7\%.

Notably, three of the top four most impactful features—\texttt{interaction\_time} (mean importance: 380.2), \texttt{propensity\_score} (341.6), and \texttt{interaction\_convenience} (339.2)—are derived directly from the SAPA framework. This demonstrates a powerful interplay between synthesized attitudes and specific trip context. Table \ref{tab:runtime_analysis} provides a summary of the computational runtime for each phase of the SAPA pipeline.

\begin{table}[h!]
\centering
\begin{tabular}{@{}lc@{}}
\toprule
\textbf{Pipeline Component} & \textbf{Approximate Runtime} \\
\midrule
LLM Persona Generation & 1-2 hours \\
LLM Latent Variable Scoring & 1-2 hours \\
Data Preparation & 5-10 minutes \\
Model Training \& Ablation Studies & 10-20 minutes \\
\bottomrule
\end{tabular}
\caption{Computational Runtime Analysis for SAPA Pipeline Components.}
\label{tab:runtime_analysis}
\end{table}

\section{Additional Validation}
\subsection{Cross-Sectional Validation of Latent Variables}
\begin{figure*}[t]
\centering
\includegraphics[width=\textwidth]{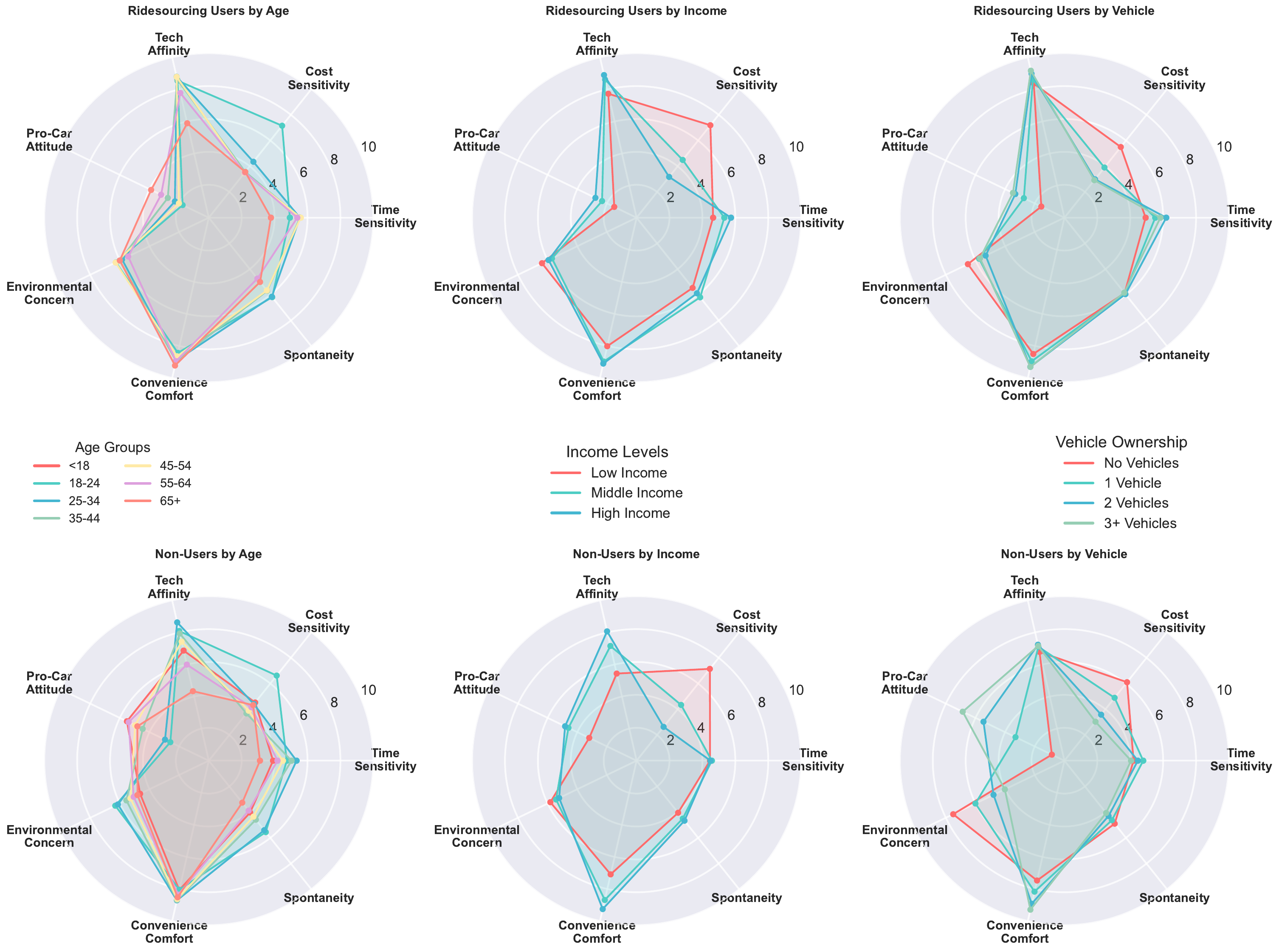}
\caption{Comparison of mean latent variable profiles for ridesourcing users (top row) and non-users (bottom row) across different demographic segments in the PSRC dataset.}
\label{fig:demographic_breakdowns}
\end{figure*}

To validate the synthesized personas, we analyzed how the mean latent variable scores differed across key demographic segments. Figure \ref{fig:demographic_breakdowns} shows the profiles for ridesourcing users and non-users, broken down by age, income, and household vehicle ownership. The distinct patterns, which align with established transportation literature, support the behavioral validity of the LLM-generated personas.

\end{appendix}

\end{document}